\documentclass{article}
\usepackage{iclr2027_conference,times}
\usepackage[T1]{fontenc}

\usepackage{amsmath,amsfonts,bm}

\def\eqref#1{equation~\ref{#1}}
\def\1{\bm{1}}

\DeclareMathAlphabet{\mathsfit}{\encodingdefault}{\sfdefault}{m}{sl}
\SetMathAlphabet{\mathsfit}{bold}{\encodingdefault}{\sfdefault}{bx}{n}

\usepackage{hyperref}
\usepackage{url}

\usepackage{booktabs}
\usepackage{multirow}
\usepackage{enumitem}
\usepackage{tcolorbox}
\tcbuselibrary{breakable}
\usepackage{xcolor,colortbl,array}
\usepackage{amssymb}

\definecolor{RevealBlue}{HTML}{7FB3D0}
\definecolor{RevealAggregate}{HTML}{DDBB78}

\newcommand{\RevealCell}[2]{%
    \cellcolor{RevealBlue!#1!white}#2%
}
\newcommand{\RevealTotal}[2]{%
    \cellcolor{RevealAggregate!#1!white}#2%
}

\definecolor{FrontierTextGray}{gray}{0.4}
\newcommand{\frontiertext}[1]{\textcolor{FrontierTextGray}{#1}}

\title{REVEAL: Reverse Engineering Agentic \\ 
Latents from Frontier-Model Traces}

\author{%
\normalfont
\textbf{Yuetian Du}\textsuperscript{1,}\thanks{Equal contribution.},
\textbf{Meng Wang}\textsuperscript{1,*},
\textbf{Yucheng Wang}\textsuperscript{1},
\textbf{Zhangzhe Chen}\textsuperscript{2},
\textbf{Wei Wang}\textsuperscript{3},
\textbf{Boyu Yang}\textsuperscript{3},\\
\normalfont
\textbf{Weiqi Zhai}\textsuperscript{3,4},
\textbf{Bing Zhao}\textsuperscript{3},
\textbf{Ming Kong}\textsuperscript{1},
\textbf{Zhan Zhou}\textsuperscript{1},
\textbf{Jie Liu}\textsuperscript{1,5,}\thanks{Corresponding authors.},
\textbf{Qiang Zhu}\textsuperscript{1,\ensuremath{\dagger}}\\
\normalfont
\textsuperscript{1}Zhejiang University\quad
\textsuperscript{2}South China University of Technology\\
\normalfont
\textsuperscript{3}Alibaba Group\quad
\textsuperscript{4}Fudan University\quad
\textsuperscript{5}City University of Hong Kong
}

\hypersetup{
  pdftitle={REVEAL: Reverse Engineering Agentic Latents from Frontier-Model Traces},
  pdfauthor={Yuetian Du, Meng Wang, Yucheng Wang, Zhangzhe Chen, Wei Wang, Boyu Yang, Weiqi Zhai, Bing Zhao, Ming Kong, Zhan Zhou, Jie Liu, Qiang Zhu}
}

\iclrpreprintcopy
\begin{document}

\maketitle

\begin{abstract}
Frontier LLM agents perform complex interactive tasks, but their inference costs motivate transferring their experience to smaller open-source agents. Distillation and latent communication typically require target-model training or access to source-model internals. We study how textual frontier-model trajectories can provide reusable local controls for a frozen agent. Inspired by protocol reverse engineering, we introduce \textit{Agentic Latents}, comprising anchored states and dynamic profiles of feedback--response cycles elicited by replay in the target model. Our training-free framework, \textbf{REVEAL} (\textbf{REV}erse \textbf{E}ngineering \textbf{A}gentic \textbf{L}atents), uses these profiles and trajectory outcomes to match and select local events from successful and failed rollouts of the same task and source model. Their response-transition contrasts yield a target-specific steering direction, applied before each response. Within-benchmark evaluations on Terminal-Bench 2.1 and Humanity's Last Exam with tools report higher accuracy than unmodified agents across three targets. Representation analyses and behavioral case studies support reusing frontier-model experience as local activation control.
\end{abstract}

\section{Introduction}
\label{sec:introduction}

Large language model (LLM) agents perform complex tasks through repeated interactions with tools and environments. Frontier models navigate these interactions effectively, but repeated calls over long trajectories can incur substantial inference costs. Smaller open-source models offer a path to lower-cost and more flexible deployment, yet often struggle with complex tool use and sustained task execution~\citep{qin2024toolllm,merrill2026terminal}. Frontier-model trajectories provide a record of how stronger agents reason, act, and respond to feedback. Reusing this experience could help smaller agents complete demanding tasks with low additional adaptation cost.

\begin{figure}
    \centering
    \includegraphics[width=0.9\linewidth]{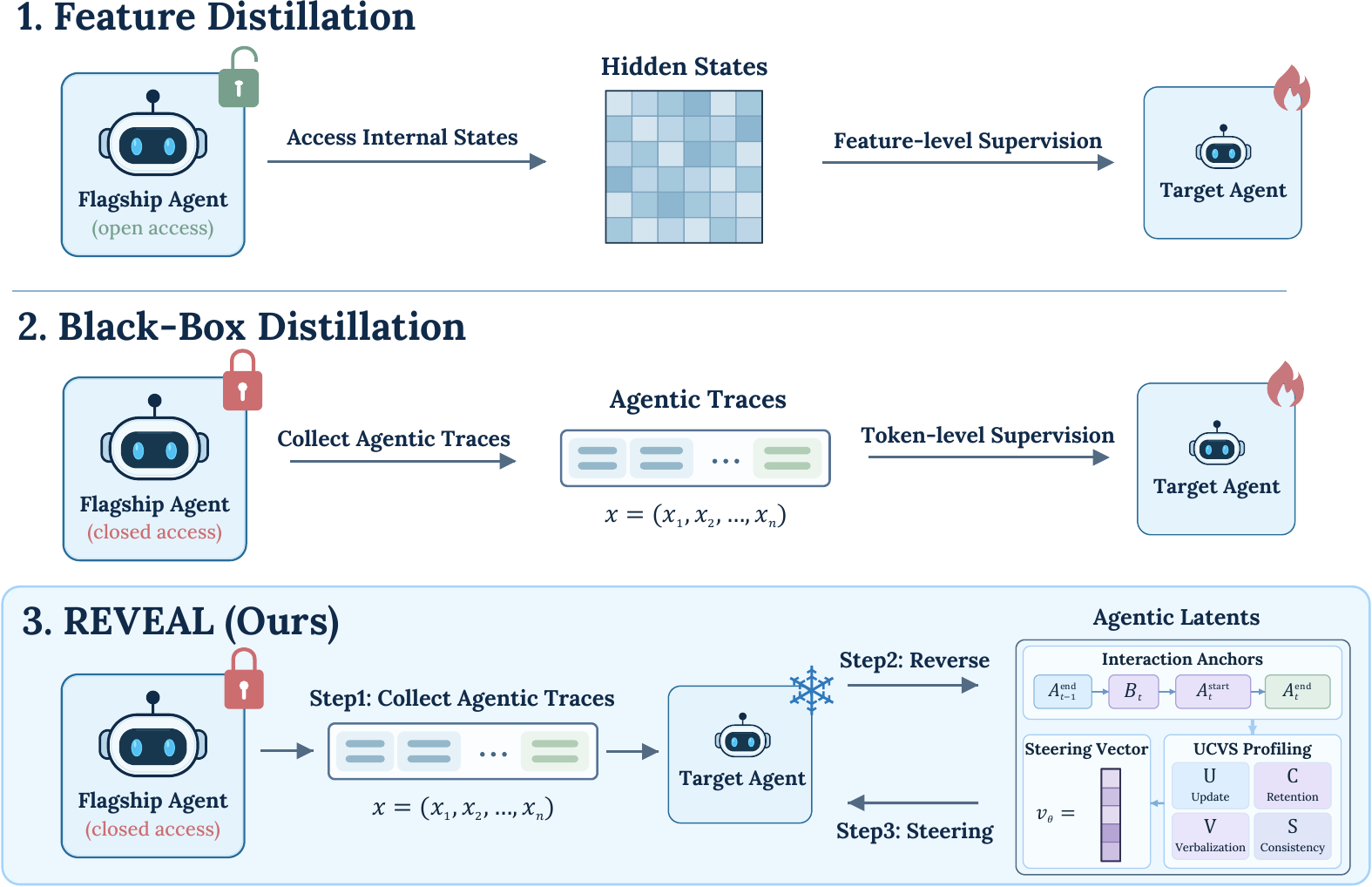}
    \caption{Three routes for reusing frontier-model experience: (1) feature distillation trains on internal states; (2) black-box distillation trains on textual trajectories; (3) REVEAL characterizes Agentic Latents through frozen target-model replay of textual trajectories and derives steering directions.}
\label{fig:transfer_comparison}
\end{figure}

Existing approaches transfer model capabilities through several forms of supervision. Teacher-generated task labels and explanations can supervise smaller models~\citep{hsieh2023distilling}, while interaction trajectories provide demonstrations of tool use and sequential decision-making~\citep{qin2024toolllm}. Access to internal states further enables representation-level transfer, including latent communication~\citep{zou2025latent,zou2026recursive}. These routes rely on target-model training or access to source-model internals. Figure~\ref{fig:transfer_comparison} contrasts the corresponding training and information-access requirements. These requirements leave a practical gap: training adds optimization costs for each target, while text-only interfaces preclude internal-state transfer. This leads us to ask: \textit{Can long-horizon text trajectories from frontier models be reconstructed as reusable local control signals for a frozen open-source agent?}

Inspired by \textit{protocol reverse engineering}, which infers message structure through comparisons of observed network traces~\citep{cui2007discoverer}, we organize frontier-model experience into local interaction units for representation analysis and control. We introduce \textit{Agentic Latents}: anchored states and dynamic profiles of feedback--response cycles, obtained by replaying interaction traces in a frozen open-source model. Representation engineering provides tools for analyzing these states and intervening on model behavior~\citep{zou2023representation}. A central challenge is that trajectory outcomes provide coarse supervision for local control: successful rollouts may contain ineffective retries, while failed ones may include useful corrections. We therefore use local representation dynamics to guide comparisons between successful and failed rollouts of the same task and frontier model.

We introduce \textbf{REVEAL} (\textbf{REV}erse \textbf{E}ngineering \textbf{A}gentic \textbf{L}atents), a training-free framework that derives activation-steering signals from Agentic Latents (Figure~\ref{fig:transfer_comparison}, bottom). Anchored states are read at corresponding stages of feedback reception and response generation across trajectories, providing a common basis for local comparison. Their dynamic profiles characterize local interaction through update magnitude ($U$), directional retention ($C$), verbal accessibility ($V$), and cross-cycle consistency ($S$), guiding event matching and selection. Each target model aggregates response-transition contrasts from the selected events into a direction in its own representation space and applies it before generating each response. 

We evaluate REVEAL on Terminal-Bench 2.1~\citep{merrill2026terminal} and Humanity's Last Exam with tools~\citep{phan2025humanity}. On Terminal-Bench, REVEAL improves task completion across all three evaluated target models; for Qwen3.6-27B, it also outperforms trajectory SFT with fewer online tokens and interactions. Complementary analyses examine how response anchors summarize local transitions, how feedback-associated directions evolve during responses, and which trajectory patterns are shared across target models. Together with behavioral case analysis, these results support using Agentic Latents to characterize long-horizon experience and derive local controls for open-source agents.

Our contributions are threefold:

\begin{itemize}[leftmargin=*, itemsep=0pt, topsep=0pt, parsep=1pt]

\item \textbf{Agentic Latents for characterizing long-horizon interaction.} We introduce Agentic Latents, comprising anchored states and dynamic profiles, to characterize local internal representations in long-horizon interaction. The states mark corresponding interaction stages, while the profiles describe within- and cross-cycle dynamics for event comparison and selection.

\item \textbf{Training-free control through trajectory reverse engineering.} We propose REVEAL, which selects matched local events from successful and failed frontier-model rollouts and constructs steering directions from their response-transition contrasts in each target model. The method uses textual trajectories and outcome labels without updating target parameters or accessing frontier-model logits or hidden states.

\item \textbf{Evidence for control utility and latent interaction structure.} We evaluate REVEAL across three target models on terminal interaction and tool-assisted reasoning. Performance and online-cost comparisons assess the utility of the reconstructed controls, while complementary analyses examine response-transition approximation, representation dynamics, cross-model consistency, and subsequent agent behavior.

\end{itemize}

\section{Method}
\label{sec:method}

\subsection{Overview}
\label{sec:method_overview}

Given textual frontier-model trajectories and task outcomes, REVEAL characterizes Agentic Latents and derives steering directions while keeping target parameters fixed. In Figure~\ref{fig:overview}(a), frozen replay yields anchored states at corresponding interaction stages and dynamic profiles of within- and cross-cycle interaction. REVEAL uses these profiles to match and select local events, then aggregates their target-space response-transition contrasts into a steering direction (Figure~\ref{fig:overview}(b)). During online interaction, the target applies this direction once before each response (Figure~\ref{fig:overview}(c)). State extraction, profile computation, and direction construction are offline; each target serves as its own observer in our experiments.

\subsection{Preliminaries}
\label{sec:preliminaries}

\noindent \textbf{Agent--environment interaction.}
Given a task $q$ and an initial observation $o_0$, the agent's history at round $t$ is $H_t=(q,o_0,a_0,o_1,\ldots,a_{t-1},o_t)$. We suppress fixed system instructions, tool specifications, and serialization in this notation. Here, $a_t$ is an assistant response, which may include tool calls, and $o_t$ is the feedback received before that response. For a nonterminal interaction,
\begin{equation}
  a_t\sim\pi_\theta(\cdot\mid H_t),
  \qquad
  o_{t+1}\sim\mathcal E(\cdot\mid H_t,a_t),
  \label{eq:interaction}
\end{equation}
where $\pi_\theta$ is the agent policy and $\mathcal E$ is the environment. A rollout $\tau$ ends with task completion or an interaction limit. An external evaluator assigns its outcome $y_\tau\in\{0,1\}$; this label is excluded from the replay input.

\noindent \textbf{Activation steering.}
Let $h^\phi_\ell(x)\in\mathbb R^{d_\phi}$ denote the residual-stream state at position $x$ and layer $\ell$ in a frozen model with parameters $\phi$; the corresponding causal token prefix is implicit. We use $\psi$ for the observer and $\theta$ for the target; the same model may serve both roles. Additive steering modifies the target state during a forward pass~\citep{panickssery2024caa}:
\begin{equation}
  h^\theta_\ell(x)
  \leftarrow
  h^\theta_\ell(x)+\alpha v^\theta_\ell,
  \label{eq:generic_steering}
\end{equation}
where $v^\theta_\ell$ is a steering direction and $\alpha$ its strength. The remaining computation uses the modified state without updating $\theta$.

\subsection{Characterizing Agentic Latents}
\label{sec:anchor_measurements}

Agentic Latents comprise two components: anchored states and dynamic profiles. The states are extracted at corresponding positions in feedback--response cycles, while the profiles characterize state changes, response-content accessibility, and cross-cycle relationships for event comparison and selection. A frozen observer replays recorded trajectories under teacher forcing with causal attention. We suppress the rollout index when describing an individual cycle.

\noindent \textbf{Anchored states.}
For each cycle $t\geq1$, we read internal states at anchors marking the preceding response endpoint, the input boundary, and the current response endpoints:
\begin{equation}
  A_{t-1}^{\mathrm{end}}
  \longrightarrow B_t
  \longrightarrow A_t^{\mathrm{start}}
  \longrightarrow A_t^{\mathrm{end}}.
  \label{eq:anchor_chain}
\end{equation}
$A_{t-1}^{\mathrm{end}}$ is the last non-whitespace content token of the preceding response. $B_t$ is the final token of the input prefix, including environmental feedback and the generation header, immediately before the current response. $A_t^{\mathrm{start}}$ and $A_t^{\mathrm{end}}$ are the first lexical and last non-whitespace content tokens of that response. Structural chat-template markers are excluded from response endpoints. Each model maps the corresponding text spans to its own token positions. Cycles with missing anchors are omitted; token-mapping conventions are given in Appendix~\ref{app:anchor_conventions}.

\begin{figure}
    \centering
    \includegraphics[width=0.95\linewidth]{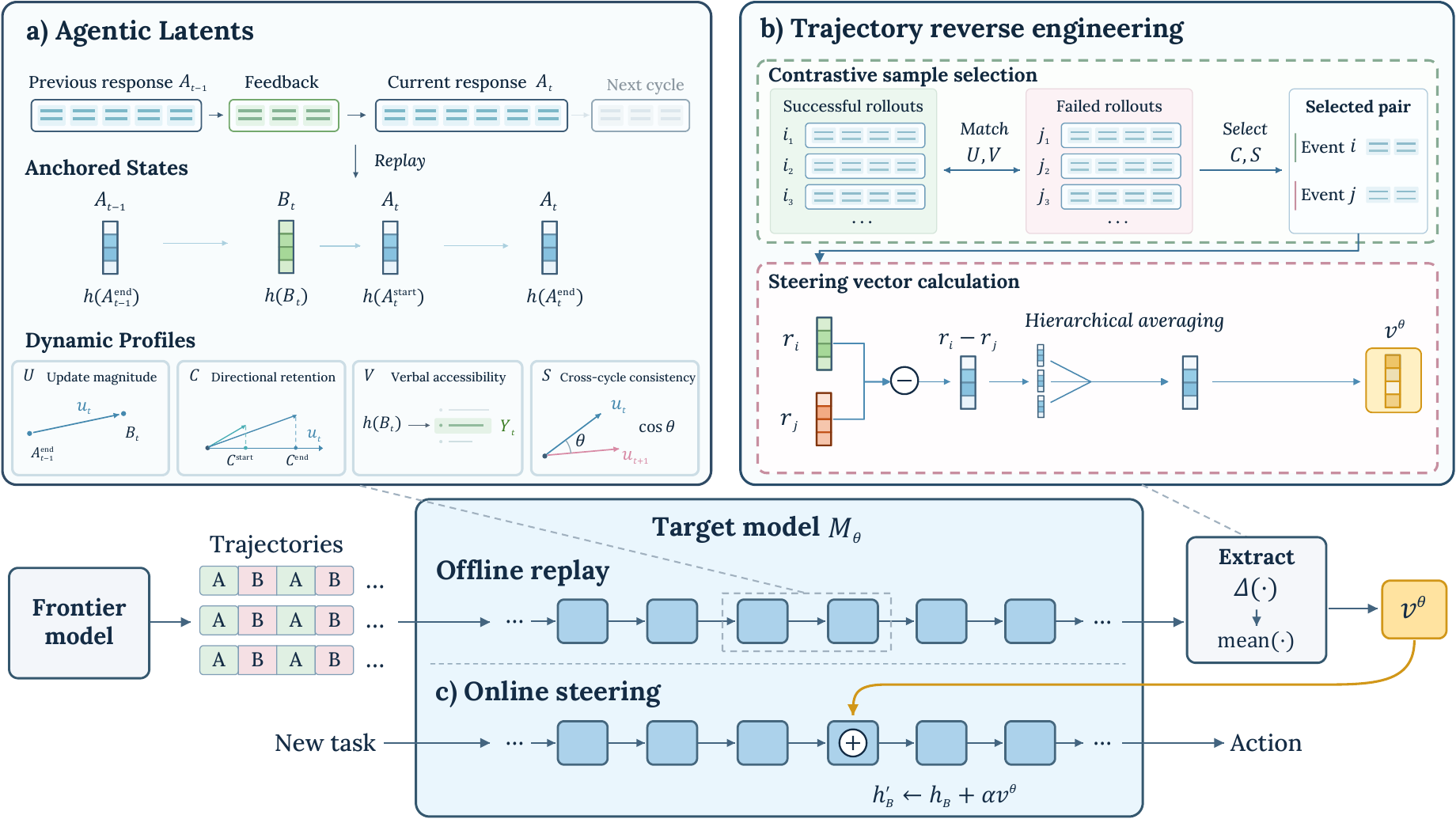}
    \caption{REVEAL reconstructs activation control from textual frontier-model trajectories. (a) Frozen replay yields Agentic Latents: anchored states and dynamic profiles. (b) Contrastive sample selection identifies matched local events; steering vector calculation combines their response-transition differences through hierarchical averaging. (c) The frozen target applies the resulting direction before each response.}
    \label{fig:overview}
\end{figure}

\noindent \textbf{Dynamic profiles.}
We characterize the anchored states through four measurement families, $U$, $C$, $V$, and $S$, for event comparison and selection. For $U$, $C$, and $S$, we independently RMS-normalize observer states as $\bar h^\psi_\ell(x)=h^\psi_\ell(x)/\sqrt{\|h^\psi_\ell(x)\|_2^2/d_\psi+\epsilon}$, where $\epsilon>0$ is a numerical stabilizer (Appendix~\ref{app:anchor_conventions}).

\noindent \textbf{Update magnitude ($U$).}
The displacement across the feedback interval and its root-mean-square magnitude are
\begin{equation}
  u_{t,\ell}
  =
  \bar h^\psi_\ell(B_t)
  -
  \bar h^\psi_\ell(A_{t-1}^{\mathrm{end}}),
  \qquad
  U_{t,\ell}
  =
  \frac{\|u_{t,\ell}\|_2}{\sqrt{d_\psi}}.
  \label{eq:update_magnitude}
\end{equation}
$U$ quantifies the representation change associated with the feedback interval.

\noindent \textbf{Directional retention ($C$).}
We project response states onto the feedback-associated update, relative to the preceding response endpoint:
\begin{equation}
  C_{t,\ell}(X)
  =
  \frac{
    \left\langle
      \bar h^\psi_\ell(X)
      -
      \bar h^\psi_\ell(A_{t-1}^{\mathrm{end}}),
      u_{t,\ell}
    \right\rangle
  }
  {\|u_{t,\ell}\|_2^2+\epsilon},
  \qquad
  X\in\{A_t^{\mathrm{start}},A_t^{\mathrm{end}}\}.
  \label{eq:directional_retention}
\end{equation}
We denote the two coefficients by $C^{\mathrm{start}}_{t,\ell}$ and $C^{\mathrm{end}}_{t,\ell}$ at the response start and end, respectively. A coefficient near one indicates a projected displacement of comparable magnitude to the original update. Values may exceed one or become negative, indicating a larger or oppositely signed component along the same direction.

\noindent \textbf{Verbal accessibility ($V$).}
We measure the accessibility of recorded response tokens using a frozen Jacobian lens (J-lens)~\citep{gurnee2026verbalizable}, with pre-fitted matrices from Neuronpedia. The boundary vocabulary readout is
\begin{equation}
  p^{J,\psi}_{t,\ell}
  =
  \operatorname{softmax}\!\left(
    W^\psi_{\mathrm U}
    \operatorname{Norm}_\psi
    \left(\widehat J^\psi_\ell h^\psi_\ell(B_t)\right)
  \right),
  \label{eq:jacobian_readout}
\end{equation}
where $\widehat J^\psi_\ell$ is a fixed residual-transport matrix for the observer, and $\operatorname{Norm}_\psi$ and $W^\psi_{\mathrm U}$ are the observer's final normalization and unembedding. Its theoretical definition is given in Appendix~\ref{app:jlens}. Let $Y_t$ contain the unique non-structural content token IDs in the recorded response, tokenized by the observer, and let $\mathcal V_\psi$ be its vocabulary. We measure their mean normalized rank:
\begin{equation}
  V_{t,\ell}
  =
  \frac{1}{|Y_t|}
  \sum_{y\in Y_t}
  \left[
    1-
    \frac{
      \operatorname{rank}_{p^{J,\psi}_{t,\ell}}(y)-1
    }
    {|\mathcal V_\psi|-1}
  \right].
  \label{eq:content_readability}
\end{equation}
Ranks are descending and one-based, and $V$ is defined for nonempty $Y_t$. Higher values place the recorded response tokens nearer the top of the boundary ranking. These tokens supply offline readout targets; causal attention prevents them from affecting the boundary state.

\noindent \textbf{Cross-cycle consistency ($S$).}
We measure the directional agreement between consecutive updates:
\begin{equation}
  S_{t,\ell}
  =
  \frac{
    \langle u_{t,\ell},u_{t+1,\ell}\rangle
  }
  {
    \|u_{t,\ell}\|_2\|u_{t+1,\ell}\|_2
  }.
  \label{eq:cross_cycle_consistency}
\end{equation}
This cosine is defined for nonzero updates in immediately adjacent cycles. Positive values indicate directional agreement, while negative values indicate disagreement. Events without a valid successor are excluded from selection.

\subsection{Trajectory Reverse Engineering and Steering}
\label{sec:reverse_engineering}

REVEAL selects contrastive event pairs using the dynamic profiles of Agentic Latents, then calculates steering directions from their anchored target states (Figure~\ref{fig:overview}(b)).

\noindent \textbf{Contrastive sample selection.}
For each task, we form pairs $p=(\tau^+,\tau^-)$ of successful and failed completed rollouts from the same frontier model, excluding infrastructure failures. Candidate samples are local feedback--response events $i=(\tau,t)$ with complete dynamic profiles, including a valid successor for $S$. In a matched pair $(i,j)$, $i$ is drawn from $\tau^+$ and $j$ from $\tau^-$.

We summarize each measurement using fixed observer-layer sets and aggregation rules, restricting $V$ to lens-supported layers. For event $i$, a summary $m_i$ is standardized as $z_m(i)=(m_i-\mu_m)/(\sigma_m+\epsilon)$ using construction-data statistics. Within each rollout pair, we match events one-to-one without replacement by their proximity in $(z_U,z_V)$, prioritizing the same tool category and similar prefix lengths. We then rank matched pairs by their differences in directional retention and cross-cycle consistency. Writing $\Delta z_m(i,j)=z_m(i)-z_m(j)$, the selection score is
\begin{equation}
  G_{ij}
  =
  \tfrac12
  \left(
    |\Delta z_{C^{\mathrm{start}}}(i,j)|
    +
    |\Delta z_{C^{\mathrm{end}}}(i,j)|
  \right)
  +
  |\Delta z_S(i,j)|.
  \label{eq:selection_score}
\end{equation}
The top half of matched pairs within each rollout pair form $\mathcal K_p$. Thus, $U$ and $V$ guide comparability, while $C$ and $S$ identify contrasting dynamics. Rollout outcomes determine the contrast's orientation; larger $C$ or $S$ values are not assumed to be better.

\begin{table*}[t]
\centering
\caption{Main agent performance (accuracy, \%). Shading is normalized within each column; darker cells indicate higher scores. Within each model block, the best and second-best \textbf{All} scores are bold and underlined, respectively.}
\label{tab:detailed_main_results}

\scriptsize
\setlength{\tabcolsep}{3pt}
\renewcommand{\arraystretch}{1.08}

\begin{tabular}{@{}l*{13}{c}@{}}
\toprule
Method & \multicolumn{4}{c}{\textbf{Terminal-Bench 2.1}} & \multicolumn{9}{c}{\textbf{HLE w/ Tools}} \\
\cmidrule(lr){2-5}\cmidrule(lr){6-14}
& Easy & Medium & Hard & \textbf{All} & Math & Phys. & Chem. & Bio/Med. & CS/AI & Eng. & Hum./Soc. & Other & \textbf{All} \\
\midrule

\multicolumn{14}{@{}l}{\strut\textbf{\textit{Qwen3.6-27B}}} \\
\midrule
Vanilla & \RevealCell{60}{75.00} & \RevealCell{79}{43.64} & \RevealCell{48}{10.00} & \RevealTotal{73}{33.71} & \RevealCell{53}{24.65} & \RevealCell{48}{14.00} & \RevealCell{60}{13.33} & \RevealCell{55}{20.93} & \RevealCell{51}{12.70} & \RevealCell{48}{15.79} & \RevealCell{39}{12.00} & \RevealCell{35}{4.44} & \RevealTotal{52}{18.00} \\
Prompt-Reg & \RevealCell{85}{100.00} & \RevealCell{69}{38.18} & \RevealCell{85}{20.00} & \RevealTotal{76}{34.83} & \RevealCell{65}{28.37} & \RevealCell{41}{12.00} & \RevealCell{60}{13.33} & \RevealCell{78}{27.91} & \RevealCell{71}{17.46} & \RevealCell{48}{15.79} & \RevealCell{45}{14.00} & \RevealCell{35}{4.44} & \RevealTotal{64}{20.80} \\
REVEAL-Base & \RevealCell{60}{75.00} & \RevealCell{76}{41.82} & \RevealCell{85}{20.00} & \RevealTotal{78}{\underline{35.96}} & \RevealCell{64}{27.91} & \RevealCell{66}{20.00} & \RevealCell{85}{20.00} & \RevealCell{55}{20.93} & \RevealCell{78}{19.05} & \RevealCell{35}{10.53} & \RevealCell{39}{12.00} & \RevealCell{60}{6.67} & \RevealTotal{65}{\underline{21.00}} \\
\textbf{REVEAL (Ours)} & \RevealCell{85}{100.00} & \RevealCell{85}{47.27} & \RevealCell{73}{16.67} & \RevealTotal{85}{\textbf{39.33}} & \RevealCell{68}{29.30} & \RevealCell{54}{16.00} & \RevealCell{85}{20.00} & \RevealCell{63}{23.26} & \RevealCell{78}{19.05} & \RevealCell{35}{10.53} & \RevealCell{50}{16.00} & \RevealCell{85}{8.89} & \RevealTotal{69}{\textbf{22.00}} \\

\addlinespace[2pt]\midrule

\multicolumn{14}{@{}l}{\strut\textbf{\textit{Gemma-4-31B-IT}}} \\
\midrule
Vanilla & \RevealCell{35}{50.00} & \RevealCell{51}{27.27} & \RevealCell{22}{3.33} & \RevealTotal{45}{20.22} & \RevealCell{73}{30.70} & \RevealCell{72}{22.00} & \RevealCell{85}{20.00} & \RevealCell{70}{25.58} & \RevealCell{58}{14.29} & \RevealCell{48}{15.79} & \RevealCell{45}{14.00} & \RevealCell{35}{4.44} & \RevealTotal{71}{22.40} \\
Prompt-Reg & \RevealCell{10}{25.00} & \RevealCell{41}{21.82} & \RevealCell{10}{0.00} & \RevealTotal{33}{14.61} & \RevealCell{65}{28.37} & \RevealCell{60}{18.00} & \RevealCell{10}{0.00} & \RevealCell{78}{27.91} & \RevealCell{65}{15.87} & \RevealCell{48}{15.79} & \RevealCell{85}{28.00} & \RevealCell{35}{4.44} & \RevealTotal{70}{22.20} \\
REVEAL-Base & \RevealCell{60}{75.00} & \RevealCell{60}{32.73} & \RevealCell{22}{3.33} & \RevealTotal{55}{\underline{24.72}} & \RevealCell{65}{28.37} & \RevealCell{85}{26.00} & \RevealCell{60}{13.33} & \RevealCell{78}{27.91} & \RevealCell{85}{20.63} & \RevealCell{60}{21.05} & \RevealCell{68}{22.00} & \RevealCell{60}{6.67} & \RevealTotal{77}{\underline{23.80}} \\
\textbf{REVEAL (Ours)} & \RevealCell{85}{100.00} & \RevealCell{57}{30.91} & \RevealCell{35}{6.67} & \RevealTotal{57}{\textbf{25.84}} & \RevealCell{85}{34.42} & \RevealCell{60}{18.00} & \RevealCell{85}{20.00} & \RevealCell{85}{30.23} & \RevealCell{71}{17.46} & \RevealCell{85}{31.58} & \RevealCell{56}{18.00} & \RevealCell{85}{8.89} & \RevealTotal{85}{\textbf{25.80}} \\

\addlinespace[2pt]\midrule

\multicolumn{14}{@{}l}{\strut\textbf{\textit{GPT-OSS-20B}}} \\
\midrule
Vanilla & \RevealCell{10}{25.00} & \RevealCell{10}{3.64} & \RevealCell{10}{0.00} & \RevealTotal{10}{3.37} & \RevealCell{15}{13.02} & \RevealCell{10}{2.00} & \RevealCell{35}{6.67} & \RevealCell{17}{9.30} & \RevealCell{10}{3.17} & \RevealCell{10}{0.00} & \RevealCell{16}{4.00} & \RevealCell{10}{2.22} & \RevealTotal{10}{7.80} \\
Prompt-Reg & \RevealCell{10}{25.00} & \RevealCell{16}{7.27} & \RevealCell{22}{3.33} & \RevealTotal{17}{\underline{6.74}} & \RevealCell{10}{11.63} & \RevealCell{10}{2.00} & \RevealCell{35}{6.67} & \RevealCell{10}{6.98} & \RevealCell{37}{9.52} & \RevealCell{22}{5.26} & \RevealCell{16}{4.00} & \RevealCell{35}{4.44} & \RevealTotal{12}{8.20} \\
REVEAL-Base & \RevealCell{10}{25.00} & \RevealCell{19}{9.09} & \RevealCell{10}{0.00} & \RevealTotal{17}{\underline{6.74}} & \RevealCell{22}{15.35} & \RevealCell{16}{4.00} & \RevealCell{60}{13.33} & \RevealCell{25}{11.63} & \RevealCell{17}{4.76} & \RevealCell{10}{0.00} & \RevealCell{10}{2.00} & \RevealCell{60}{6.67} & \RevealTotal{18}{\underline{9.80}} \\
\textbf{REVEAL (Ours)} & \RevealCell{35}{50.00} & \RevealCell{22}{10.91} & \RevealCell{22}{3.33} & \RevealTotal{24}{\textbf{10.11}} & \RevealCell{34}{19.07} & \RevealCell{16}{4.00} & \RevealCell{35}{6.67} & \RevealCell{32}{13.95} & \RevealCell{30}{7.94} & \RevealCell{35}{10.53} & \RevealCell{16}{4.00} & \RevealCell{10}{2.22} & \RevealTotal{28}{\textbf{12.00}} \\

\bottomrule
\end{tabular}

\vspace{3pt}
\parbox{\textwidth}{\scriptsize
\textit{Terminal-Bench protocols.} Terminal-Bench 2.1 includes 4 easy, 55 medium, and 30 hard tasks. \\
\textit{HLE categories.} Math, Phys., Chem., Bio/Med., CS/AI, Eng., and Hum./Soc. denote Mathematics, Physics, Chemistry, Biology/Medicine, Computer Science/AI, Engineering, and Humanities/Social Science. 
}
\end{table*}

\noindent \textbf{Steering vector calculation.}
Frozen target replay provides the raw anchor states for each selected event and its recorded prefix. For $i=(\tau,t)$, we denote its anchors by $B_i$, $A_i^{\mathrm{start}}$, and $A_i^{\mathrm{end}}$, and compute the response-transition summary
\begin{equation}
  r^\theta_{\ell,i}
  =
  \tfrac12
  \left[
    h^\theta_\ell(A_i^{\mathrm{start}})
    +
    h^\theta_\ell(A_i^{\mathrm{end}})
  \right]
  -
  h^\theta_\ell(B_i).
  \label{eq:response_transition}
\end{equation}
This summary averages the response-start and response-end displacements from the input boundary. Each selected pair $(i,j)\in\mathcal K_p$ contributes $r^\theta_{\ell,i}-r^\theta_{\ell,j}$, contrasting boundary-relative response displacements from successful and failed source rollouts in the same target layer. \textit{Hierarchical averaging} averages these contrasts within each rollout pair, then averages pair means equally within each task and task means equally across tasks (Appendix~\ref{app:hierarchical_aggregation}). We normalize the resulting nonzero aggregate as $v^\theta_\ell=\widetilde v^\theta_\ell/\|\widetilde v^\theta_\ell\|_2$ and use it as a candidate direction for perturbing the target's pre-response state. The recorded response endpoints are required only during offline construction.

\noindent \textbf{Online steering.}
For each target and benchmark, we use a fixed intervention layer $\ell_s$ and strength $\alpha$, chosen empirically. Before each response, we apply
\begin{equation}
  h^\theta_{\ell_s}(B_t)
  \leftarrow
  h^\theta_{\ell_s}(B_t)
  +
  \alpha v^\theta_{\ell_s}.
  \label{eq:reveal_steering}
\end{equation}
The addition occurs once during prefill at the final prefix token; $B_0$ denotes this position for the initial response. Decoding then proceeds normally. The direction and strength remain fixed, and online execution requires no additional observer pass or frontier-model query.

\section{Experiments}
\label{sec:experiments}

\subsection{Tasks and Datasets}
\label{sec:tasks_datasets}

We evaluate REVEAL on terminal interaction and tool-assisted reasoning. \textbf{Terminal-Bench 2.1}~\citep{merrill2026terminal} comprises 89 tasks that require agents to operate in containerized command-line environments. We use the complete task set and evaluate task completion with the benchmark's task-specific verifiers through Harbor.\footnote{\url{https://github.com/harbor-framework/terminal-bench-2-1}} Results are reported overall and by the official difficulty categories: 4 easy, 55 medium, and 30 hard tasks. \textbf{Humanity's Last Exam (HLE)}~\citep{phan2025humanity} evaluates expert-level reasoning across academic disciplines. We use the fixed 500-question subset released by WebThinker~\citep{li2025webthinker}, which was randomly sampled from the text-only questions, and evaluate it in a tool-assisted setting. Final answers are graded by GPT-5.6-luna; we report accuracy overall and across eight subject categories.

\subsection{Baselines}
\label{sec:baselines}

We compare REVEAL with three baselines. \textbf{Vanilla} uses the underlying agent without additional feedback-regulation instructions or activation steering. \textbf{Prompt-Reg} adds a fixed natural-language policy that instructs the agent to revise assumptions using feedback, carry updates into subsequent actions, avoid ineffective repetition, and preserve valid constraints. The full policy appears in Appendix~\ref{app:prompt_reg}. \textbf{REVEAL-Base} constructs a steering direction from target-model activations elicited by successful and failed frontier-model trajectories, adapting contrastive activation addition (CAA)~\citep{panickssery2024caa}. It uses trajectory outcomes without REVEAL's profile-guided selection of local feedback--response events. For the efficiency comparison, we also fine-tune Qwen3.6-27B by next-token prediction on the same 124 frontier-model trajectories used for direction construction.

\begin{table*}[t]
\centering
\caption{Accuracy and online interaction costs on Terminal-Bench 2.1. The row in gray reports the frontier model's mean pass@1 over four rollouts per task; the remaining rows use Qwen3.6-27B. Token and interaction costs are means per task. Bold values indicate the best results among the three target-model methods.}
\label{tab:efficiency}
\small
\setlength{\tabcolsep}{2.5pt}
\renewcommand{\arraystretch}{1.13}
\begin{tabular}{@{}lccccccccc@{}}
\toprule
& \multicolumn{4}{c}{\textbf{Accuracy (\%)} $\uparrow$}
& \multicolumn{2}{c}{\textbf{Online tokens (k)} $\downarrow$}
& \multicolumn{2}{c}{\textbf{Interactions} $\downarrow$}
& \shortstack{\textbf{Target}\\\textbf{training}} \\
\cmidrule(lr){2-5}\cmidrule(lr){6-7}\cmidrule(lr){8-9}
Method & Easy & Medium & Hard & All
& Input & Output & Rounds & Tool calls & \\
\midrule
\frontiertext{Frontier Model}
& \frontiertext{93.75} & \frontiertext{60.00}
& \frontiertext{39.17} & \frontiertext{54.49}
& \frontiertext{346.16} & \frontiertext{9.46}
& \frontiertext{16.03} & \frontiertext{28.56}
& \frontiertext{--} \\
\midrule
Vanilla
& 75.00 & 43.64 & 10.00 & 33.71
& 595.14 & 15.21 & 23.30 & 47.91 & No \\
Vanilla + SFT
& 75.00 & \textbf{49.09} & 6.67 & 35.96
& 223.14 & 14.02 & 15.70 & 34.87 & Yes \\
\textbf{REVEAL}
& \textbf{100.00} & 47.27 & \textbf{16.67} & \textbf{39.33}
& \textbf{143.71} & \textbf{6.95}
& \textbf{13.88} & \textbf{27.92} & No \\
\bottomrule
\end{tabular}
\end{table*}

\subsection{Implementation Details}
\label{sec:implementation_details}

We use Qwen3.6-27B~\citep{qwen3.6-27b}, Gemma-4-31B-IT~\citep{gemmateam2026gemma4}, and GPT-OSS-20B~\citep{openai2025gptoss120bgptoss20bmodel} as target models, each serving as its own frozen observer. GPT-5.6-sol generates four rollouts per task. Tasks with both successful and failed rollouts contribute selected contrasts to one steering direction per target and benchmark. We evaluate within-benchmark adaptation using outcome-labeled source trajectories from the evaluation task collection, without a held-out task split. Target parameters remain frozen.

We choose the intervention layer and strength empirically and keep them fixed during evaluation. On Terminal-Bench, agents use Terminus-2 through Harbor, with vLLM serving at temperature 1.0. REVEAL intervenes at layers 40, 30, and 15 for Qwen3.6-27B, Gemma-4-31B-IT, and GPT-OSS-20B, respectively, with $\alpha=0.5$, at the final input-prefix position before each response. Per-request generation caps are 81,920, 8,192, and 32,768 tokens, respectively. We set these caps separately based on the models' observed response-generation behavior during agent interactions, and share each cap across methods for the same target. They bound individual responses; actual task-level consumption is measured separately in the efficiency analysis. 

\begin{figure*}[t]
\centering
\includegraphics[width=0.9\textwidth]{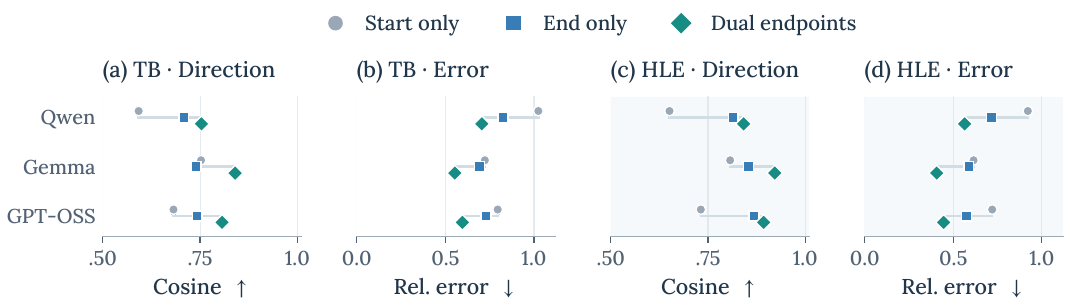}
\caption{Endpoint approximations of mean response displacement across three models and two benchmarks; points and bars show task-balanced means and 95\% task-bootstrap intervals. Layers 23, 22, and 8 are used for Qwen, Gemma, and GPT-OSS, respectively.}
\label{fig:anchor_fidelity}
\end{figure*}

\begin{figure*}[t]
\centering
\includegraphics[width=0.85\textwidth]{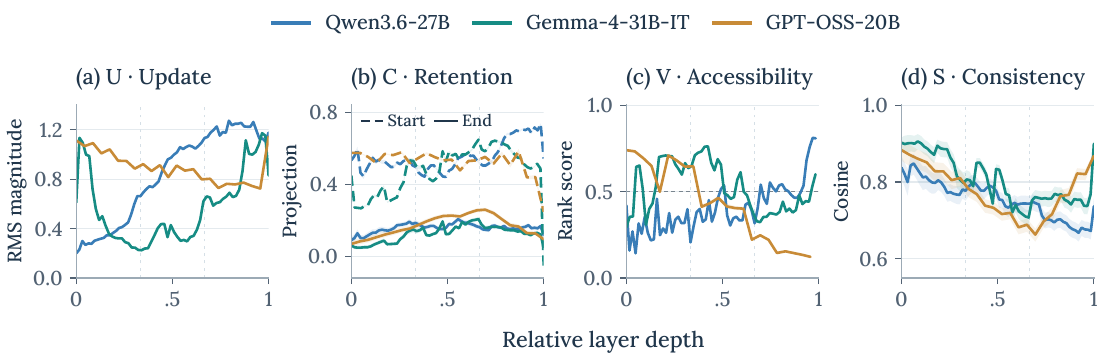}
\caption{Layerwise representation profiles on Terminal-Bench with pointwise 95\% task-bootstrap intervals; dashed and solid $C$ curves denote response-start and response-end projections.}
\label{fig:representation_profiles}
\end{figure*}

\subsection{Main Results}
\label{sec:main_results}

\textbf{Finding 1: REVEAL achieves higher reported performance across both benchmarks and all three target models.} On Terminal-Bench, completion improves by 5.62--6.74 percentage points over Vanilla and by 1.12--3.37 points over REVEAL-Base (Table~\ref{tab:detailed_main_results}). On HLE, REVEAL likewise achieves the highest reported overall accuracy among the compared methods for each target model.

\label{sec:efficiency}

\textbf{Finding 2: REVEAL reduces online interaction costs relative to trajectory SFT.} Using the same 124 trajectories from 33 tasks within the evaluation collection, REVEAL solves 35/89 Terminal-Bench tasks versus 32/89 for trajectory SFT (Table~\ref{tab:efficiency}). Relative to SFT, it uses 35.60\% fewer input tokens and 50.43\% fewer output tokens, with fewer rounds and tool calls. These measurements cover execution after adaptation; source collection, offline replay, profile computation, and SFT training are excluded.

\subsection{Empirical Analysis}
\label{sec:empirical_analysis}

\textbf{Finding 3: Response anchors provide a compact summary of local representation transitions.} At the measured layer of each model and on both benchmarks, averaging the response-start and response-end states better approximates the boundary-relative mean response displacement than either endpoint alone (Figure~\ref{fig:anchor_fidelity}). Mean cosine similarity reaches $0.75$--$0.92$, while relative error decreases by $14.5$--$30.9\%$ against the better single endpoint. These results support the response-endpoint summary used to construct steering contrasts from Agentic Latents.

\begin{figure*}[t]
\centering
\includegraphics[width=0.87\textwidth]{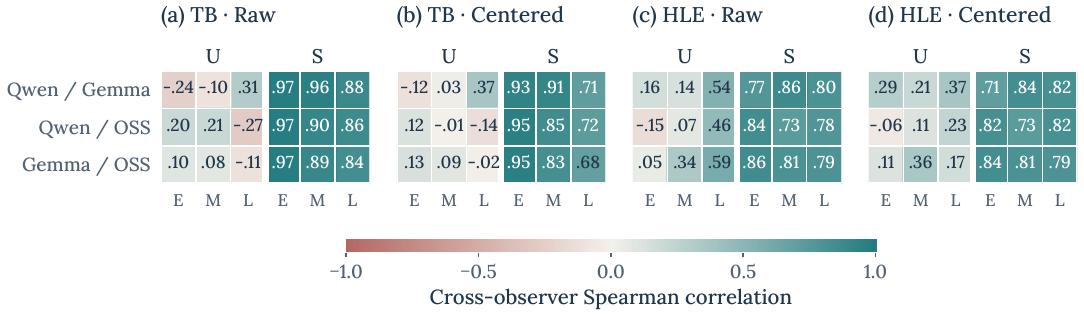}
\caption{Cross-model Spearman correlations of trajectory-level $U$ and $S$ across depth bands, before and after task centering on matched trajectories. E/M/L denote early/middle/late bands.}
\label{fig:cross_observer_structure}
\end{figure*}

\begin{figure}[!tbp]
    \centering
    \includegraphics[width=0.9\linewidth]{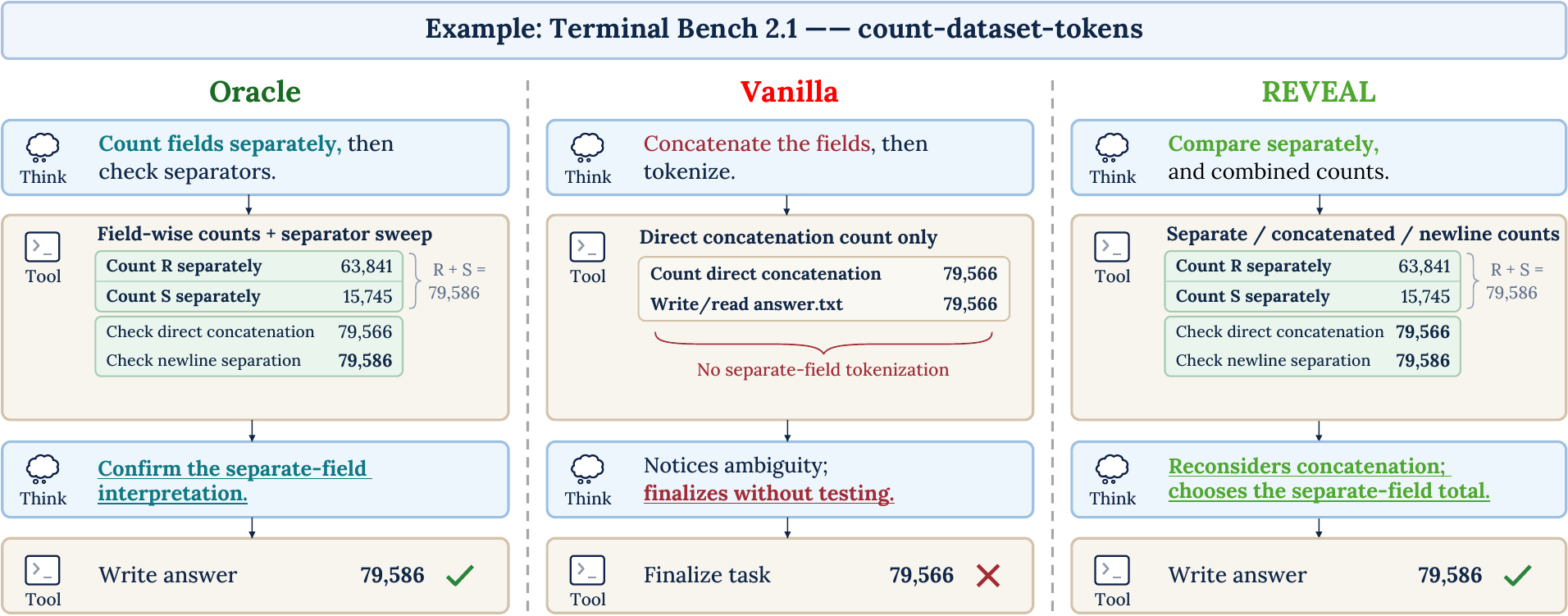}
    \caption{A case study. REVEAL and the frontier reference compare alternative token-counting operations, whereas Vanilla finalizes without testing the recognized ambiguity.}
    \label{fig:case_study}
\end{figure}

\textbf{Finding 4: Update magnitude and directional retention reveal complementary interaction dynamics.} Response-end projections are lower than response-start projections in 147/148 evaluated layers on Terminal-Bench and 144/148 on HLE. Figure~\ref{fig:representation_profiles} places this shared pattern alongside the distinct depth profiles of update magnitude, verbal accessibility, and cross-cycle consistency. Full HLE profiles and within-response curves extend this analysis in Appendices~\ref{app:hle_profiles}--\ref{app:response_retention}. 

\textbf{Finding 5: The same trajectories induce correlated cross-cycle dynamics across models.} For identical replayed trajectories, task-centered $S$ profiles retain cross-model Spearman correlations of $0.68$--$0.95$ on Terminal-Bench and $0.71$--$0.84$ on HLE (Figure~\ref{fig:cross_observer_structure}). Agreement in $U$ is substantially weaker: shared temporal structure coexists with model-dependent update magnitudes. Outcome-conditioned contrasts retain this agreement (Appendix~\ref{app:outcome_contrasts}). These correlations characterize scalar trajectory descriptors; steering directions are reconstructed separately in each target's latent space.

\textbf{Finding 6: In a counting task, the steered agent adopts the frontier reference's verification pattern.} In \texttt{count-dataset-tokens}, Vanilla recognizes a possible separator effect but finalizes the direct-concatenation total of 79,566 without testing alternatives. The frontier reference counts fields independently and checks separator variants. REVEAL uses a different counting script, compares the alternatives, and revises its initial preference for concatenation before selecting 79,586 (Figure~\ref{fig:case_study}). This case illustrates a verification pattern shared with the frontier reference, expressed through the target's own computation and decision sequence.

\section{Related Work}
\label{sec:related-work}

\subsection{Representation Engineering and Activation Steering}

Representation engineering examines how high-level concepts are encoded in LLM representations and how interventions on these representations alter model behavior \citep{bartoszcze2025representation,wehner2025taxonomy}. The Representation Engineering framework \citep{zou2023representation} systematizes this perspective through two complementary operations: Representation Reading and Representation Control. Representative studies probe truth-related representations through Inference-Time Intervention \citep{li2023iti} and The Geometry of Truth \citep{marks2024geometry}, while Activation Addition \citep{turner2023steering} and Contrastive Activation Addition \citep{panickssery2024caa} construct steering directions from prompt-pair or contrastive activation differences. Related vector-based studies examine refusal \citep{arditi2024refusal}, in-context functions \citep{todd2024functionvectors}, and task representations \citep{hendel2023taskvectors}, while evaluation work shows that measured steering effects depend on protocol design \citep{pres2024evaluation}. REVEAL compares feedback--response dynamics across trajectories through the anchored states and dynamic profiles of Agentic Latents.

\subsection{Tool Use and Long-Horizon Agents}

Research on tool-using LLM agents increasingly examines multi-step interactions shaped by observations, tool outputs, and environmental feedback \citep{qin2024tool,wang2024survey}. Representative systems include WebGPT \citep{nakano2022webgpt} for browsing-grounded question answering, ReAct \citep{yao2023react} for reasoning--action interleaving, Toolformer \citep{schick2023toolformer} and ToolLLM \citep{qin2024toolllm} for API use, and Reflexion \citep{shinn2023reflexion} for feedback-based retry. Recent work reads tool necessity from pre-generation states and controls it through textual prefill, diagnoses capability-relative knowing--doing gaps, and extracts steerable tool-identity directions \citep{sun2026when2tool,cheng2026modeladaptive,wu2026toolcalling}. Trajectory-level methods further decode tool dependencies and predict episode failure from agent states, or audit safety risks through observer representations \citep{sun2026dependency,ruan2026doomed,li2026traces}. Closer to long-horizon control, From Actions to Understanding \citep{padhi2026actions} derives success/failure directions over agent trajectories, whereas TACT \citep{sui2026tact} steers behavioral drift axes at reasoning--action boundaries. REVEAL uses trajectory outcomes and local representation profiles to select interaction events, then derives steering directions from their target-model response transitions.

\section{Conclusion}
\label{sec:conclusion}

Long-horizon experience can be reused as local control within an agent's existing representation space. REVEAL represents this experience as \textit{Agentic Latents}, comprising anchored states and dynamic profiles, and uses the profiles to select response-transition contrasts for steering. Our results connect these controls to improved task execution and, on Terminal-Bench, more economical online interaction. This reuses frontier-model experience through the executing model's own representations, without accessing frontier-model internals or updating target parameters.

\section*{Limitations}
Our evaluations measure within-benchmark adaptation using outcome-labeled trajectories from the evaluation task collection; transfer to held-out tasks and new environments remains untested. Future work could evaluate explicit construction/evaluation splits and transfer across frontier sources. Event selection also depends on compatible pre-fitted Jacobian lenses and valid feedback--response anchors, excluding events with missing measurements. Alternative readouts and selection rules for incomplete interaction records could extend the settings in which Agentic Latents can be studied, while testing whether their control utility persists.

\subsection*{AI use statement}
We used generative AI tools to assist with literature and citation checks, local editing to improve the clarity and readability of the manuscript, and LaTeX template migration and formatting. The authors reviewed and revised the outputs as needed at each stage and take full responsibility for the final text, claims, citations, and artifacts.

\bibliography{references}
\bibliographystyle{iclr2027_conference}

\clearpage

\appendix

\section{Jacobian-Lens Readout}
\label{app:jlens}

The verbal-accessibility component of Agentic Latents compares a boundary-state vocabulary readout with the content of the recorded response. This section develops the Jacobian-lens construction underlying that readout and distinguishes its role in event selection from the construction of REVEAL's steering direction.

\subsection{From Local Perturbations to a Shared Readout}

Following \citet{gurnee2026verbalizable}, consider the effect of perturbing a source residual state on downstream residual states. For a fixed calibration context $c$, let $z^\psi(x';c)$ denote the residual state at the lens's output layer, and define
\begin{equation}
 D^\psi_{\ell;x',x}(c)
 =\frac{\partial z^\psi(x';c)}{\partial h^\psi_\ell(x;c)}
 \in\mathbb R^{d_\psi\times d_\psi}.
 \label{eq:jlens_local_jacobian}
\end{equation}
Write $z^\psi=h^\psi_{\ell_{\mathrm{out}}}$, where $\ell_{\mathrm{out}}$ is the output-layer index of the fitted lens; the published construction considers final and penultimate output layers. A perturbation $\delta$ at source position $x$ gives the first-order expansion
\begin{equation}
 z^{\psi,\delta}(x';c)-z^\psi(x';c)
 =D^\psi_{\ell;x',x}(c)\delta+o(\|\delta\|_2),
 \qquad x'\geq x.
 \label{eq:jlens_perturbation}
\end{equation}
Causal attention makes the derivative zero for $x'<x$. The derivatives are evaluated on a fixed token sequence, so the expansion describes a forward computation without differentiating through future-token sampling.

For a final-layer target, the corresponding logit derivative follows by the chain rule. Writing $s^\psi_{\mathrm{out}}(x';c)=W^\psi_{\mathrm U}\operatorname{Norm}_\psi(z^\psi(x';c))$,
\begin{equation}
 \frac{\partial s^\psi_{\mathrm{out}}(x';c)}{\partial h^\psi_\ell(x;c)}
 =W^\psi_{\mathrm U}\,
 D\operatorname{Norm}_\psi\bigl(z^\psi(x';c)\bigr)
 D^\psi_{\ell;x',x}(c).
 \label{eq:jlens_logit_derivative}
\end{equation}
This identity separates residual transport from the output head. In particular, the normalization derivative depends on the context; averaging residual Jacobians is not the same operation as averaging logit derivatives.

Let $\mathcal I_c$ and $\mathcal O_c$ be the valid source- and destination-position sets for context $c$. Matching the official mean estimator\footnote{\url{https://github.com/anthropics/jacobian-lens}}, we define the population lens by summing over causal destinations and averaging over source positions and contexts:
\begin{equation}
 J^\psi_\ell
 =\mathbb E_c\!\left[
   \frac{1}{|\mathcal I_c|}\sum_{x\in\mathcal I_c}
   \sum_{\substack{x'\in\mathcal O_c\\x'\geq x}}
   D^\psi_{\ell;x',x}(c)\right].
 \label{eq:jacobian_average}
\end{equation}
For a calibration corpus $\mathcal D$, the corresponding empirical estimator is
\begin{equation}
 \widehat J^\psi_\ell
 =\frac{1}{|\mathcal D|}
   \sum_{c\in\mathcal D}\frac{1}{|\mathcal I_c|}
   \sum_{x\in\mathcal I_c}
   \sum_{\substack{x'\in\mathcal O_c\\x'\geq x}}
   D^\psi_{\ell;x',x}(c).
 \label{eq:jlens_estimator}
\end{equation}
Here $\mathcal D$ is the lens-calibration corpus. The destination sum should not be replaced by a uniform mean over position pairs: that change generally reweights source positions. Source layers, output layer, position masks, and aggregation choices are properties of the fitted checkpoint.

REVEAL loads the observer-compatible matrices published by Neuronpedia\footnote{\url{https://huggingface.co/neuronpedia/jacobian-lens}}, which were fitted using the official library. We do not estimate Jacobians or refit a lens on REVEAL trajectories. In our evaluated setup, each target also serves as its own observer, so its $V$ measurements use its corresponding fitted lens. Lens fitting is an external prerequisite; REVEAL's trajectory processing and online steering keep both the model and lens parameters fixed.

\subsection{Vocabulary Scores and the J-Space}

For the raw boundary state $h=h^\psi_\ell(B_t)$, the readout first transports $h$, then applies the observer's output normalization and unembedding:
\begin{equation}
 s^\psi_{t,\ell}
 =W^\psi_{\mathrm U}\operatorname{Norm}_\psi
     (\widehat J^\psi_\ell h),
 \qquad p^{J,\psi}_{t,\ell}=\operatorname{softmax}(s^\psi_{t,\ell}).
 \label{eq:jlens_scores_appendix}
\end{equation}
The transported activation is a readout surrogate, not the reconstructed final state of this particular response. REVEAL uses the score ordering without treating the scores as calibrated probabilities of subsequent token occurrence.

The role of normalization can be made explicit for bias-free RMS normalization. If $\operatorname{Norm}_\psi(z)=\Gamma_\psi z/\sqrt{d_\psi^{-1}\|z\|_2^2+\epsilon_{\mathrm N}}$, with diagonal gain $\Gamma_\psi$, then
\begin{equation}
 s^\psi_{t,\ell}(y)
 =\frac{\langle q^\psi_{\ell,y},h\rangle}
 {\sqrt{d_\psi^{-1}\|\widehat J^\psi_\ell h\|_2^2+\epsilon_{\mathrm N}}},
 \qquad
 q^\psi_{\ell,y}
 =(\widehat J^\psi_\ell)^\top\Gamma_\psi^\top
   (W^\psi_{\mathrm U})^\top e_y,
 \label{eq:jlens_effective_token_direction}
\end{equation}
where $e_y$ is the vocabulary unit vector. The denominator is shared across tokens, so their order is determined by the inner products. This derivation requires the stated normalization form; Equation~(\ref{eq:jlens_scores_appendix}) retains the actual model head. Geometric RMS normalization in Appendix~\ref{app:anchor_conventions} is a separate operation and is not substituted into this readout.

\citet{gurnee2026verbalizable} describe the J-space through sparse nonnegative combinations of token-indexed lens directions. Using the effective token directions above to incorporate the normalization gain, and suppressing the observer superscript, let $N=|\mathcal V_\psi|$ and $Q_\ell=[q_{\ell,1},\ldots,q_{\ell,N}]$. For a sparsity budget $k$, the analogous construction is
\begin{equation}
 \mathcal J_{\ell,k}
 =\{Q_\ell\beta:\beta\geq0,\ \|\beta\|_0\leq k\},
 \qquad
 \beta^\star\in\mathop{\arg\min}_{\beta\geq0,\ \|\beta\|_0\leq k}
       \|h-Q_\ell\beta\|_2^2.
 \label{eq:jlens_sparse_cones}
\end{equation}
This is a union of cones; the sparse representation need not be unique. REVEAL computes vocabulary ranks directly and does not solve this sparse reconstruction problem. Its Agentic Latents comprise anchored states and dynamic profiles, not a recovered J-space, and its steering directions come from raw target-model response-transition contrasts rather than token-lens vectors.

\subsection{Why This Readout Fits REVEAL}

REVEAL needs a descriptor of response-content accessibility before a response begins. A logit lens reads intermediate states with the output head directly, equivalent to the identity transport in Equation~(\ref{eq:jlens_scores_appendix}). This omits the transformation between the source layer and the output representation. A tuned lens learns affine maps to approximate the model's final prediction distribution~\citep{belrose2023tunedlens}. That objective addresses next-token prediction fidelity, rather than the accessibility of content distributed throughout a response. Here, $Y_t$ includes content throughout a recorded response, and $V$ asks how highly that content ranks at the boundary. The boundary state uses its causal prefix; $Y_t$ is obtained retrospectively solely for offline scoring. We therefore use the averaged-transport readout to characterize this broader content set.

The lens operates only during offline profiling. Its $V$ score is paired with $U$ to match events with similar update size and content accessibility; $C$ and $S$ then identify contrasting dynamics. The resulting event indices determine which raw response transitions contribute to steering. Thus the lens affects the choice of evidence used to construct a direction, without defining that direction's coordinates or adding a lens computation to online execution.

\subsection{Interpreting the Empirical Readout}

Our readout diagnostics distinguish relative rank from top-token recovery. Alongside $V$, we compute
\begin{equation}
 \operatorname{Recall@50}_{t,\ell}
 =\frac{1}{|Y_t|}\sum_{y\in Y_t}
   \mathbf 1\{\operatorname{rank}_{p^{J,\psi}_{t,\ell}}(y)\leq50\}.
 \label{eq:jlens_recall50}
\end{equation}
Both statistics weight unique content token IDs equally. In task-balanced layer-band summaries of our HLE source-trajectory cohort (1,303 tasks), Qwen has late-layer $V=0.635$ with Recall@50 of $1.4089\%$, while GPT-OSS has early-layer $V=0.710$ with Recall@50 of $0.0245\%$. The respective supported bands are zero-based layers 43--62 and 0--7. This replay cohort is separate from the 500-question online evaluation. Above-chance average ranks can therefore coexist with low top-50 coverage. This supports interpreting $V$ as relative accessibility of recorded content, rather than recovery of the complete response. It also motivates using $V$ within an observer and layer convention instead of treating its absolute value as a model-independent quality score. These diagnostics characterize the chosen readout; they do not establish comparative superiority over alternative lenses.

\section{Agentic Latents: Anchors and Dynamic Profiles}
\label{app:anchor_conventions}

Agentic Latents combine states at corresponding stages of interaction with profiles of their local dynamics. The anchors define which events can be compared; the profiles specify what is compared. We give the extraction conventions and derive two properties that explain the information retained by these measurements.

\subsection{Anchored States and Causal Prefixes}

For each model, anchors are mapped from response-content and generation-header spans in its chat serialization. $A_{t-1}^{\mathrm{end}}$ is the last non-whitespace content token of the preceding response; $B_t$ is the final generation-header token after feedback; $A_t^{\mathrm{start}}$ and $A_t^{\mathrm{end}}$ are the first lexical and last non-whitespace content tokens of the current response. Structural chat markers are excluded from response endpoints. Correspondence refers to interaction roles, not equal token indices or shared hidden-state coordinates across models.

Each anchor is evaluated with its recorded causal prefix. In particular, $h^\psi_\ell(B_t)$ includes feedback and the generation header but no token of the current response. The state at $A_t^{\mathrm{start}}$ already includes the first lexical response token. Teacher forcing makes the recorded response available for extracting its later states and content set $Y_t$, while causal attention prevents those later tokens from changing the boundary state. A one-token response may have coincident start and end anchors.

The interval from $A_{t-1}^{\mathrm{end}}$ to $B_t$ includes new feedback, serialization, and a change of token position. Its displacement is therefore feedback-associated, not an isolated causal effect of the feedback text. The response anchors delimit the subsequent state change using the same reference point. Their endpoint average is a compact response summary whose fidelity is evaluated in Figure~\ref{fig:anchor_fidelity}.

\subsection{Normalization and Geometric Interpretation}

For $U$, $C$, and $S$, we independently RMS-normalize observer states without learned gains:
\begin{equation}
 \bar h^\psi_\ell(x)
 =\frac{h^\psi_\ell(x)}
 {\sqrt{d_\psi^{-1}\|h^\psi_\ell(x)\|_2^2+\epsilon}},
 \qquad \epsilon>0.
 \label{eq:observer_normalization}
\end{equation}
This gives $\|\bar h\|_2\leq\sqrt{d_\psi}$ and hence $0\leq U_{t,\ell}\leq2$. It controls state scale without identifying coordinates across models. The J-lens reads raw boundary states, and steering directions are constructed from raw target states; neither operation uses Equation~(\ref{eq:observer_normalization}).

For nonzero raw states $a_0,b_0$ at the preceding endpoint and boundary, let $a,b$ be their normalized states and $\rho_a=\|a\|_2/\sqrt{d_\psi}$, $\rho_b=\|b\|_2/\sqrt{d_\psi}$. Expanding the squared displacement gives
\begin{equation}
 U^2=\rho_a^2+\rho_b^2
      -2\rho_a\rho_b\cos(a_0,b_0).
 \label{eq:normalized_update_geometry}
\end{equation}
When the stabilizer is negligible relative to both raw mean squares, $\rho_a,\rho_b\approx1$ and $U^2\approx2[1-\cos(a_0,b_0)]$. Thus $U$ measures geometric change, not the usefulness of new information. The following decomposition explains why it is accompanied by retention and cross-cycle measurements.

\noindent\textbf{Proposition 1 (Complementary within- and cross-cycle geometry).}
Fix a layer $\ell$ and suppress its index in $U$, $C$, $S$, and $u$. Let $u_t\neq0$, $x_t^{\mathrm{pre}}=\bar h^\psi_\ell(A_{t-1}^{\mathrm{end}})$, and $w_t(X)=\bar h^\psi_\ell(X)-x_t^{\mathrm{pre}}$ for a response anchor $X$. Then
\begin{equation}
 w_t(X)
 =\left(1+\frac{\epsilon}{d_\psi U_t^2}\right)C_t(X)u_t
   +w_t^\perp(X),
 \qquad \langle w_t^\perp(X),u_t\rangle=0.
 \label{eq:profile_projection_decomposition}
\end{equation}
For a nonzero update in the immediately succeeding cycle,
\begin{equation}
 \frac{\|u_{t+1}-u_t\|_2^2}{d_\psi}
 =U_{t+1}^2+U_t^2-2U_tU_{t+1}S_t.
 \label{eq:profile_temporal_decomposition}
\end{equation}
Equal update magnitudes can therefore accompany different response retention and cross-cycle directions.

\noindent\textit{Proof.}
The orthogonal projection coefficient of $w_t(X)$ onto $u_t$ is $\langle w_t(X),u_t\rangle/\|u_t\|_2^2$. Substituting $C_t(X)=\langle w_t(X),u_t\rangle/(\|u_t\|_2^2+\epsilon)$ and $\|u_t\|_2^2=d_\psi U_t^2$ gives Equation~(\ref{eq:profile_projection_decomposition}). Expanding the squared difference between adjacent updates gives Equation~(\ref{eq:profile_temporal_decomposition}). For an explicit example with $d_\psi\geq2$, choose orthonormal $e_1,e_2$ and $0<R<\sqrt{d_\psi}$. Set $x_t^{\mathrm{pre}}=Re_1$ and $\bar h^\psi_\ell(B_t)=Re_2$. If the response endpoint returns to $Re_1$ and the next boundary is $Re_2$, then $C_t^{\mathrm{end}}=0$ and $S_t=1$. If the response endpoint remains at $Re_2$ and the next boundary returns to $Re_1$, then $C_t^{\mathrm{end}}=2R^2/(2R^2+\epsilon)$ and $S_t=-1$. Both cases have $U_t=U_{t+1}=R\sqrt{2/d_\psi}$. The chosen normalized radius is attainable under Equation~(\ref{eq:observer_normalization}) by raw vectors of norm $R\sqrt{\epsilon/(1-R^2/d_\psi)}$. These are examples of admissible measurement geometry, not claims about which continuations a particular model produces.

The decomposition distinguishes retention along the update axis from changes in orthogonal directions. In particular,
\begin{equation}
 C_t^{\mathrm{end}}-C_t^{\mathrm{start}}
 =\frac{\left\langle
 \bar h^\psi_\ell(A_t^{\mathrm{end}})-\bar h^\psi_\ell(A_t^{\mathrm{start}}),u_t
 \right\rangle}{\|u_t\|_2^2+\epsilon}.
 \label{eq:retention_response_change}
\end{equation}
A decrease indicates displacement against that axis during the response, without implying that all feedback information has disappeared. At equal update magnitudes, Equation~(\ref{eq:profile_temporal_decomposition}) becomes $2U_t^2(1-S_t)$: $S$ distinguishes temporal directions even when update sizes are unchanged. These properties explain why the empirical depth profiles need not move together.

\subsection{Verbal Accessibility as a Rank Statistic}

The response supplies a set $Y_t$ of unique non-structural content token IDs. Using a set avoids weighting a token repeatedly because it occurs many times. Let $N=|\mathcal V_\psi|$ and $m=|Y_t|$. The score in Equation~(\ref{eq:content_readability}) averages how many vocabulary entries rank below each response token. This admits a precise pairwise interpretation.

\noindent\textbf{Proposition 2 (Ranking interpretation and chance reference).}
Assume $N>1$ and $1\leq m<N$, and consider a strict vocabulary ranking with fixed tie-breaking. Let $\operatorname{AUC}_{t,\ell}$ be the fraction of pairs $(y,z)\in Y_t\times(\mathcal V_\psi\setminus Y_t)$ in which $y$ ranks above $z$. Then
\begin{equation}
 V_{t,\ell}
 =\frac{(N-m)\operatorname{AUC}_{t,\ell}+(m-1)/2}{N-1}.
 \label{eq:rank_auc_relation}
\end{equation}
Any strictly increasing transformation of the scores preserves $V$. Under a uniformly random vocabulary permutation independent of $Y_t$,
\begin{equation}
 \mathbb E[V_{t,\ell}]=\tfrac12,
 \qquad
 \operatorname{Var}(V_{t,\ell})
 =\frac{(N+1)(N-m)}{12m(N-1)^2}.
 \label{eq:rank_null_moments}
\end{equation}

\noindent\textit{Proof.}
For each $y$, $N-\operatorname{rank}(y)$ counts entries below it. Summing over $Y_t$ counts $m(N-m)\operatorname{AUC}_{t,\ell}$ comparisons with its complement and exactly $m(m-1)/2$ comparisons within $Y_t$. Dividing by $m(N-1)$ gives Equation~(\ref{eq:rank_auc_relation}). Strictly increasing transformations preserve all pairwise orders. Under the stated null, the $m$ ranks are sampled without replacement from $\{1,\ldots,N\}$. Their normalized mean has expectation $1/2$. The population variance of $(N-r)/(N-1)$ is $(N+1)/[12(N-1)]$; multiplying by the finite-population factor $(N-m)/[m(N-1)]$ gives the stated variance.

This proposition shows that $V$ is invariant to a common positive score temperature and has a shared random-order reference. It does not imply that real token ranks are uniformly distributed: token frequency and task vocabulary can produce systematic preferences. Moreover, $m$ and $N$ affect the attainable range,
\begin{equation}
 \frac{m-1}{2(N-1)}\leq V_{t,\ell}
 \leq1-\frac{m-1}{2(N-1)}.
 \label{eq:rank_attainable_range}
\end{equation}
If $m=N$, $V=1/2$ identically. We therefore retain the observer and layer context when interpreting this relative-accessibility score. Neither an above-chance $V$ nor a high top-$k$ recall assigns correctness to a response.

\subsection{From Profiles to Event Comparisons}

The profiles serve different roles in REVEAL. Matching on standardized $U$ and $V$ seeks events with comparable update size and boundary accessibility of response content. Selection on absolute differences in $C$ and $S$ then identifies contrasting within-response and cross-cycle dynamics. The final rollout outcomes orient each contrast; large $C$ or $S$ is not intrinsically rewarded. A negative $S$, for example, can accompany a correction that reverses an earlier update.

The two propositions justify interpreting these quantities separately and establish what their numerical values measure. They do not identify the correctness of an intermediate action or remove every contextual difference between matched events. The matched raw target-state transitions supply the steering direction, whose behavioral utility is evaluated by online intervention. This keeps the descriptive role of Agentic Latents distinct from the experimental test of their use for control.

\subsection{Hierarchical Averaging of Response-Transition Contrasts}
\label{app:hierarchical_aggregation}

For each task $q$, let $\mathcal P_q$ denote the retained successful--failed rollout pairs with nonempty selected event sets $\mathcal K_p$, and let $\mathcal Q=\{q:\mathcal P_q\neq\varnothing\}$ denote the eligible tasks. For each $(i,j)\in\mathcal K_p$, event $i$ comes from the successful rollout and event $j$ from the failed rollout. Their response-transition summaries are computed from raw target-model states using Equation~(\ref{eq:response_transition}).

For nonempty $\mathcal Q$, hierarchical averaging gives
\begin{equation}
  \widetilde v^\theta_\ell
  =
  \frac{1}{|\mathcal Q|}
  \sum_{q\in\mathcal Q}
  \frac{1}{|\mathcal P_q|}
  \sum_{p\in\mathcal P_q}
  \frac{1}{|\mathcal K_p|}
  \sum_{(i,j)\in\mathcal K_p}
  \left(
    r^\theta_{\ell,i}
    -
    r^\theta_{\ell,j}
  \right).
  \label{eq:hierarchical_aggregation}
\end{equation}

Thus, selected event pairs receive equal weight within each rollout pair, rollout-pair means receive equal weight within each task, and task means receive equal weight across tasks. This prevents tasks with more eligible rollout pairs, or rollout pairs with more selected events, from receiving greater total weight solely because of their counts. Individual contrasts and intermediate means are not normalized; only the final nonzero aggregate is L2-normalized to obtain $v^\theta_\ell$.

\subsection{Validity and Missing Measurements}

Missing anchors are not interpolated. Near-zero updates are excluded from directional comparisons using the implementation's numerical thresholds. Evaluating the same projection expression at $B_t$ gives $\|u_{t,\ell}\|_2^2/(\|u_{t,\ell}\|_2^2+\epsilon)$, rather than exactly one. When $\|u_{t,\ell}\|_2^2$ is small, this shrinkage and the numerically unstable direction motivate excluding the event rather than interpreting its retention coefficient.

$S$ uses immediately adjacent cycles in the same trajectory. A missing successor is not assigned zero, and missing cycles are not skipped to create an artificial pair. Events with an empty response-content set or unavailable lens readout are excluded from selection; $V$ is aggregated only over supported lens layers. These rules distinguish unavailable measurements from genuine zero values and preserve the temporal unit used in the comparisons.

\section{Prompt-Reg Policy}
\label{app:prompt_reg}

Prompt-Reg supplies the following fixed policy as additional agent instructions. It expresses feedback-related behavioral guidance in natural language without computing representation profiles or modifying activations.

\begin{tcolorbox}[title={Prompt-Reg policy},breakable,colback=gray!3,colframe=black!55,boxrule=0.5pt,arc=1mm,left=6pt,right=6pt,top=6pt,bottom=6pt,fontupper=\small,fonttitle=\bfseries]
\textbf{Feedback regulation policy:}

After every tool or environment response, incorporate the new evidence into the current task state and revise any assumption that it contradicts. Carry the resulting update into the next decision and action. When feedback shows that the current approach is ineffective, change the relevant part of the approach rather than repeating it without new justification.

Prefer actions that directly reduce the remaining uncertainty or make verifiable progress. Reuse evidence already obtained, and avoid repeating checks or waiting for additional observations unless they are needed to decide the next action.

Preserve valid constraints and accumulated evidence, as well as the task's required response and tool-call format. Do not output this policy or an explicit reflection unless the task requires one.
\end{tcolorbox}

\section{Additional Analyses of Agentic Latents}
\label{app:additional_latent_analyses}

We examine response-summary fidelity, profile dynamics, and outcome-conditioned contrasts using the existing source-trajectory records: 89 Terminal-Bench tasks and 1,303 HLE questions, distinct from the 500-question online evaluation. Full-history replay measures at most 12 systematically spaced centers per trajectory, retaining adjacent cycles for $S$. Missing measurements are excluded. We average events within trajectories and trajectories within tasks, weighting tasks equally. Pointwise task-bootstrap intervals use 1,000 resamples for the saved profiles and 2,000 for paired analyses (seed 0), conditional on the recorded trajectories and centers.

\subsection{Anchor Fidelity Across Response Lengths}
\label{app:anchor_length}

\begin{figure}[!htb]
\centering
\includegraphics[width=\textwidth]{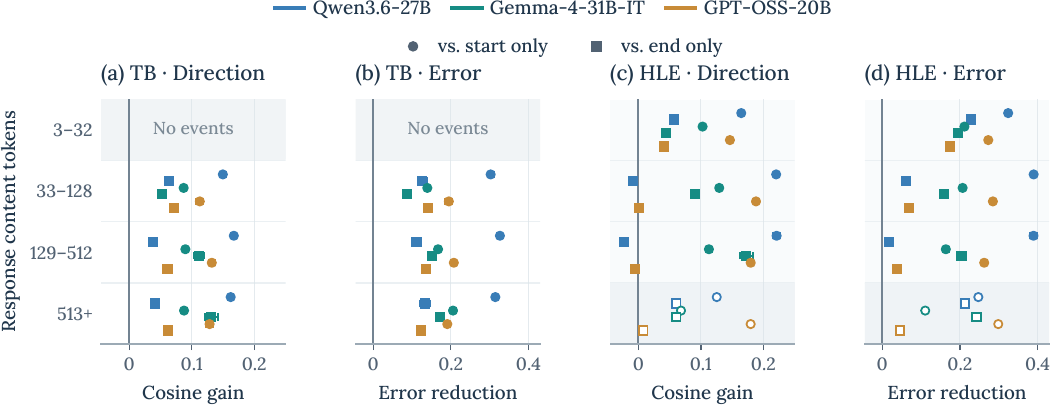}
\caption{Dual endpoints versus individual response anchors, stratified by response length. Positive values indicate higher cosine similarity or lower relative error against the full-response mean displacement. Bars show paired 95\% task-bootstrap intervals; hollow markers denote strata with fewer than three tasks.}
\label{fig:appendix_anchor_length}
\end{figure}

\noindent\textbf{Dual endpoints reduce reconstruction error across response lengths.} We stratify anchor fidelity by response-content length at layers 23, 22, and 8 for Qwen3.6-27B, Gemma-4-31B-IT, and GPT-OSS-20B, respectively (zero-indexed). Figure~\ref{fig:appendix_anchor_length} compares the endpoint average and single anchors with the boundary-relative full-response mean on the same events. Mean relative error decreases in all 42 comparisons across 21 nonempty model--benchmark--length strata. Mean cosine similarity improves in 39 of these comparisons; the exceptions favor the end anchor for Qwen at 33--128 and 129--512 tokens and GPT-OSS at 129--512 tokens on HLE. Thus, the compact response-transition summary reduces reconstruction error across lengths, while its directional advantage depends on the stratum.

\subsection{Representation Dynamics on HLE}
\label{app:hle_profiles}

\begin{figure}[!htb]
\centering
\includegraphics[width=\textwidth]{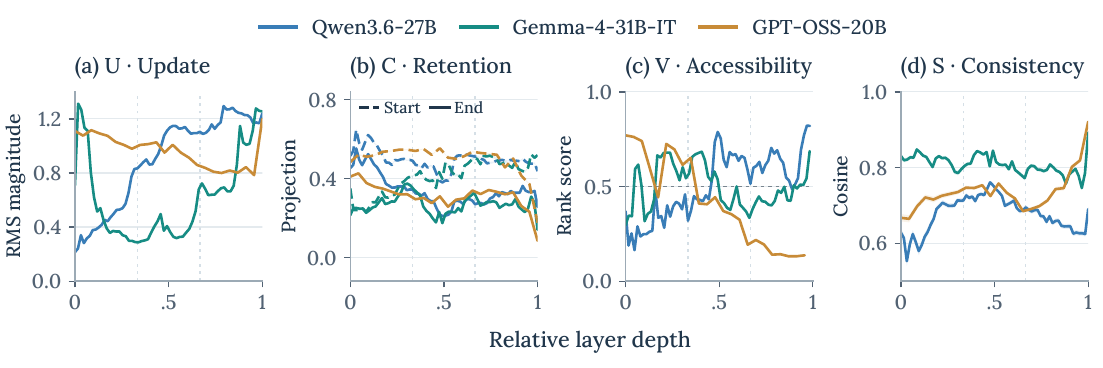}
\caption{Full-depth HLE profiles with pointwise 95\% task-bootstrap intervals. Dashed and solid $C$ curves mark response start and end, respectively. $U/C/V$ use 1,303 tasks; $S$ uses 1,091 with valid adjacent updates. Only horizontal depth is rescaled; $V$ excludes unsupported final layers.}
\label{fig:appendix_hle_profiles}
\end{figure}

\noindent\textbf{Lower response-end retention persists on HLE despite different update profiles.} Figure~\ref{fig:appendix_hle_profiles} extends the full-depth analysis to tool-assisted reasoning. Qwen's update magnitude generally increases with depth, Gemma exhibits a middle-layer trough, and GPT-OSS's update magnitude declines before a final-layer rise. Despite these differences, response-end retention is below response-start retention in 144/148 layers; the four exceptions occur in Gemma's early layers. Verbal accessibility and cross-cycle consistency also follow distinct depth profiles. Together, the profiles distinguish update size, retention, accessibility, and consistency when comparing events.

\subsection{Retention Within a Response}
\label{app:response_retention}

\begin{figure}[!htb]
\centering
\includegraphics[width=\textwidth]{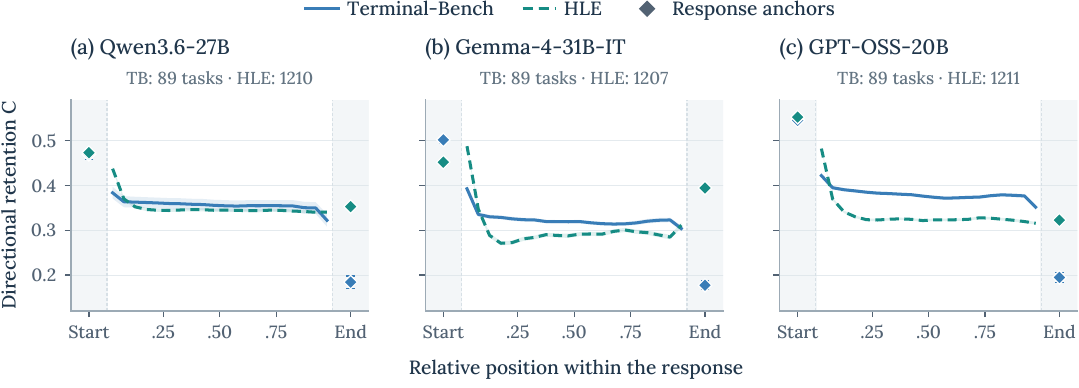}
\caption{Within-response retention at layers 23/22/8 for Qwen/Gemma/GPT-OSS. Lines show 20 relative-position bin means; separate diamonds mark the response-start and response-end anchors. All recorded events with at least 20 content tokens are included. Shading and bars show 95\% task-bootstrap intervals.}
\label{fig:response_retention}
\end{figure}

\noindent\textbf{The same endpoint ordering can arise from different response dynamics.} At the diagnostic layers used above, mean end-anchor retention is below start-anchor retention in all six model--benchmark groups (Figure~\ref{fig:response_retention}). The intermediate paths nevertheless differ. On Terminal-Bench, retention at the final token is $0.126$--$0.157$ below the last-bin mean; on HLE, it is $0.007$--$0.083$ above it. Gemma shows the clearest HLE recovery, reaching $C^{\mathrm{end}}=0.394$ after lower middle-response projections. Endpoint attenuation therefore does not imply uniform decay throughout generation. Distinct anchors preserve these stage-specific changes when local response transitions are compared.

The length criterion avoids empty bins and keeps the set of events identical across positions: 2,935/2,935/2,923 Terminal-Bench events and 9,524/9,062/9,254 HLE events for Qwen/Gemma/GPT-OSS. Single-token anchors remain distinct from bin means.

\subsection{Readout Rank and Content Coverage}
\label{app:readout_calibration}

\begin{figure}[!htb]
\centering
\includegraphics[width=\textwidth]{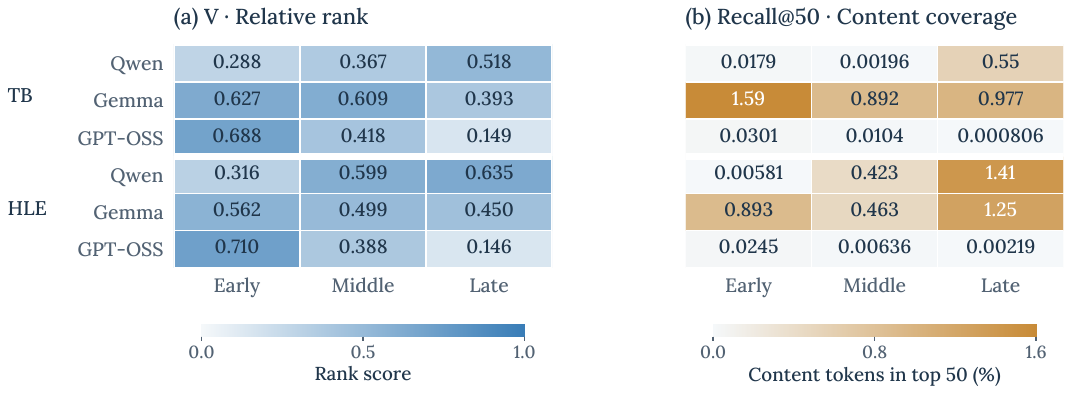}
\caption{Task-balanced $V$ and Recall@50 across early/middle/late depth bands on Terminal-Bench and HLE, using supported J-lens layers. Panels use separate linear scales; uniformly random vocabulary ranking has expected $V=0.5$.}
\label{fig:appendix_readout_calibration}
\end{figure}

\noindent\textbf{High relative content rank need not imply concentrated content recovery.} Figure~\ref{fig:appendix_readout_calibration} compares $V$ with Recall@50, the fraction of unique response-content tokens among the top 50 vocabulary entries in the boundary J-lens readout. GPT-OSS reaches $V=0.710$ in early HLE layers, but Recall@50 is only $0.0245\%$; the largest band-level recall across all groups is $1.59\%$. Relative vocabulary rank and top-ranked content coverage thus capture different properties. In REVEAL, $V$ describes accessibility for matching events within each observer, rather than action correctness or recovery of the full response.

\begin{figure}[!htb]
\centering
\includegraphics[width=\textwidth]{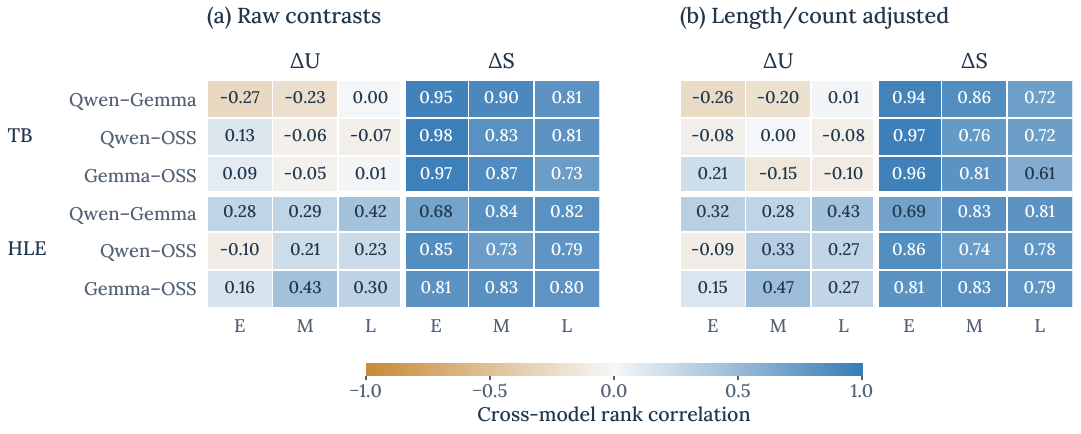}
\caption{Cross-model correlations of same-task success--failure contrasts before and after adjustment for response length and analysis-center count. E/M/L denote early/middle/late depth bands. Every panel uses the same 34 Terminal-Bench and 208 HLE tasks; $\Delta U$ provides the update-magnitude comparison.}
\label{fig:outcome_contrasts}
\end{figure}

Throughout the band-level analyses, early/middle/late denote zero-based layers 0--21/22--42/43--63 for Qwen, 0--19/20--39/40--59 for Gemma, and 0--7/8--15/16--23 for GPT-OSS. Readout summaries exclude each model's final layer, where the fitted lens is unavailable.

\subsection{Outcome-Conditioned Structure Across Observers}
\label{app:outcome_contrasts}

\noindent\textbf{Success--failure contrasts in cross-cycle consistency remain aligned across observers.} For each task with successful and failed frontier-model rollouts, we subtract the two groups' mean profiles and correlate these taskwise contrasts across models. Across 34 Terminal-Bench and 208 HLE tasks, $\Delta S$ correlations range from $0.729$ to $0.977$ and from $0.684$ to $0.855$, respectively (Figure~\ref{fig:outcome_contrasts}). After rank-based adjustment for response length and analysis-center count, they remain $0.609$--$0.971$ and $0.688$--$0.856$. The sign of $\Delta S$ varies across tasks: successful rollouts do not uniformly have higher consistency. This supports comparing task-specific dynamics by contrast magnitude, as in REVEAL's selection score, while using outcomes to orient the resulting contrasts.

\noindent\textbf{Paired analysis.} We use known-outcome source trajectories with complete three-band $U/S$ profiles, average trajectories within each task/outcome class, and retain tasks shared by all observers. Adjustment regresses ranked profile contrasts on an intercept and four ranked covariates: each observer's success--failure differences in mean $\log(1+\text{response characters})$ and $\log(1+\text{analysis centers})$. Character counts are trajectory means before transformation. Correlations use the residual ranks; bootstrap resamples refit ranks and regressions. All 18 adjusted $\Delta S$ intervals remain above zero. These task-level associations complement local-event selection; they do not establish its causal contribution.
\end{document}